\documentclass[letterpaper, 10 pt, conference]{ieeeconf}  
\usepackage{graphicx}
\usepackage{float}
\usepackage{multicol}

\IEEEoverridecommandlockouts                              

\title{\LARGE \bf
HARMONIC: A Framework for Explanatory Cognitive Robots
}

\author{Sanjay Oruganti$^{1}$, Sergei Nirenburg$^{1}$, Marjorie McShane$^{1}$, Jesse English$^{1}$,\\ Michael Roberts$^{1}$ and Christian Arndt$^{1}$
\thanks{$^{1}$Authors are with the Cognitive Science Department at Rensselaer Polytechnic Institute, Troy, NY, USA. \tt\small e-mail: sanjayovs@ieee.org}
}

\begin{document}

\maketitle
\thispagestyle{empty}
\pagestyle{empty}

\begin{abstract}

We present HARMONIC, a framework for implementing cognitive robots that transforms general-purpose robots into trusted teammates capable of complex decision-making, natural communication and human-level explanation. The framework supports interoperability between a strategic (cognitive) layer for high-level decision-making and a tactical (robot) layer for low-level control and execution. We describe the core features of the framework and our initial implementation, in which HARMONIC was deployed on a simulated UGV and  drone involved in a multi-robot search and retrieval task.

\end{abstract}

\section{INTRODUCTION}

While today’s general-purpose robots are rapidly advancing towards achieving dexterous capabilities comparable to human workers \cite{rossini2023real, malik2024intelligent, atkeson2015no}, they still face several significant limitations. They lack cognitive abilities to assess the semantics of situations, states, and actions of both themselves and others, or engage in meaning-oriented, human-level dialog. They also struggle to deal with disturbances and novel situations, which limits their adaptability in dynamic environments.  Finally, they fall short of being trusted because they are unable to generate causal explanations of their reasoning and action.   

To serve as trusted partners, robots must be able to: reliably collaborate on complex and novel tasks and missions; interpret and anticipate teammates' actions and needs; communicate with teammates in natural language; learn new skills on the job from language and demonstration; explain their own and others' decisions and actions; and teach teammates through language and demonstration.



The HARMONIC framework we present aims to meet these challenges by facilitating the implementation of embodied robots that can remember, plan, reason, explain, negotiate, learn, and teach. Specifically, the  architecture enables robots to perform:

\begin{enumerate}
    \item physical actions, such as repairing, cleaning and gofering;
    \item the mental actions needed to emulate human-like behavior, such as meaning-oriented language processing; reasoning about plans, goals, and attitudes; explaining the reasons for their own and others' actions; and accessing and archiving institutional memory (which is key to operating in complex environments); and
    \item hybrid actions, such as teaching and learning physical and mental actions through natural language and visual demonstration. 
\end{enumerate}

 The HARMONIC framework (Figure \ref{fig:overview}) is an extension of hybrid control systems and architectures as summarized by Dennis et al. \cite{dennis2016practical} and is an enhancement over the type 2 integration in the DIARC framework \cite{scheutz2013systematic, schermerhorn2006diarc}, in which concurrent and dynamic operation is facilitated by incorporating the strategic layer as a subsystem within the tactical layer. In HARMONIC, by contrast, the strategic and tactical layer components function independently and interactively. Moreover, while we implemented HARMONIC using specific cognitive and robotic control systems, this framework was designed to facilitate implementation using any state-of-the-art cognitive and robotic system that supports all the required functionalities.

While basic strategic layer functionality can be implemented using generalist models \cite{padalkar2023open}, including Large Language Models (LLMs) and Vision-Language-Action (VLA) models \cite{brohan2023rt}, these  cannot support human-level explainability, which is crucial for establishing trust with humans \cite{kambhampati2024llms}. To ensure explainability, VLAs and LLMs are limited to specific modules and functionalities within the HARMONIC framework. For details of such integration, see \cite{oruganti2023automating} and \cite{mcshane2024agents}.

\section{The Framework}
\begin{figure}[t]
    \centering
    \includegraphics[width=0.49\textwidth]{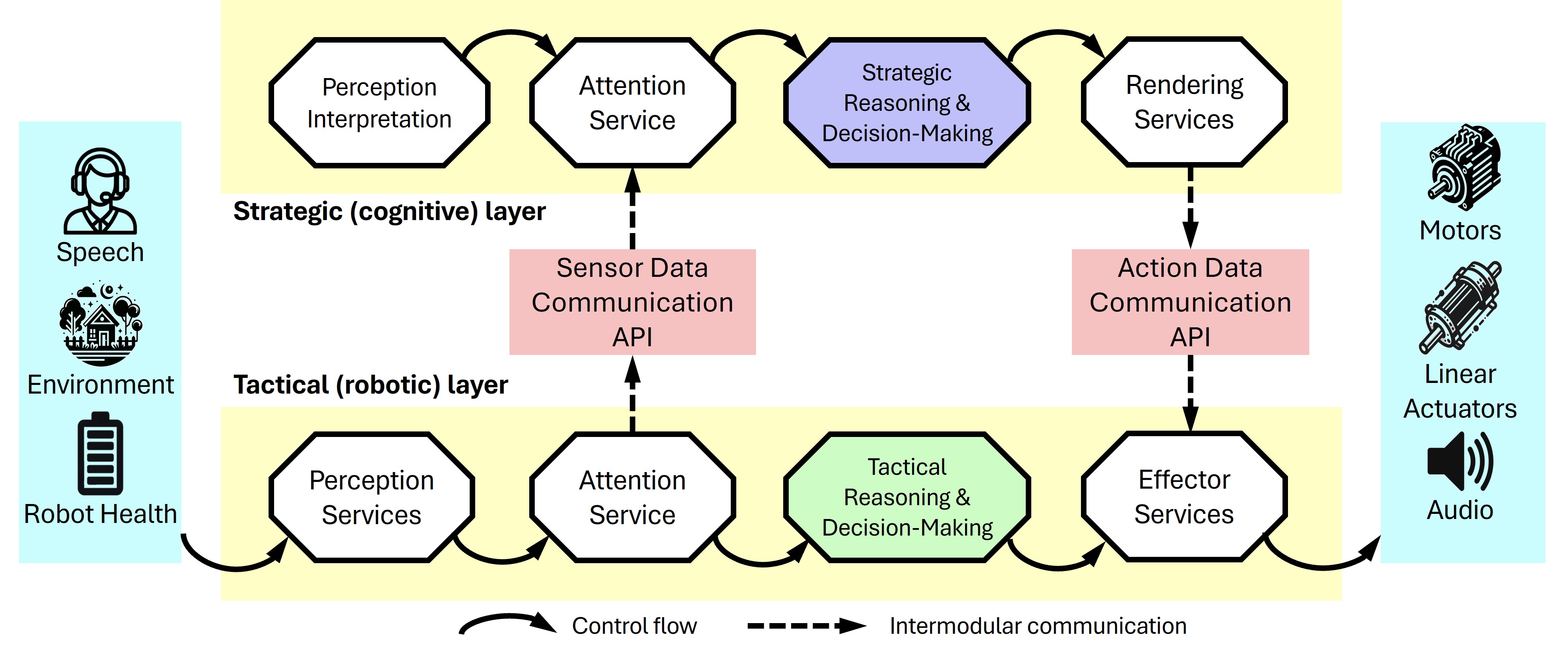}
    \caption{An overview of the HARMONIC framework showing the strategic and tactical layers.}
    \label{fig:overview}
\vspace{-17pt}
\end{figure}

\begin{figure*}[t]
    \centering
    \includegraphics[width=\textwidth]{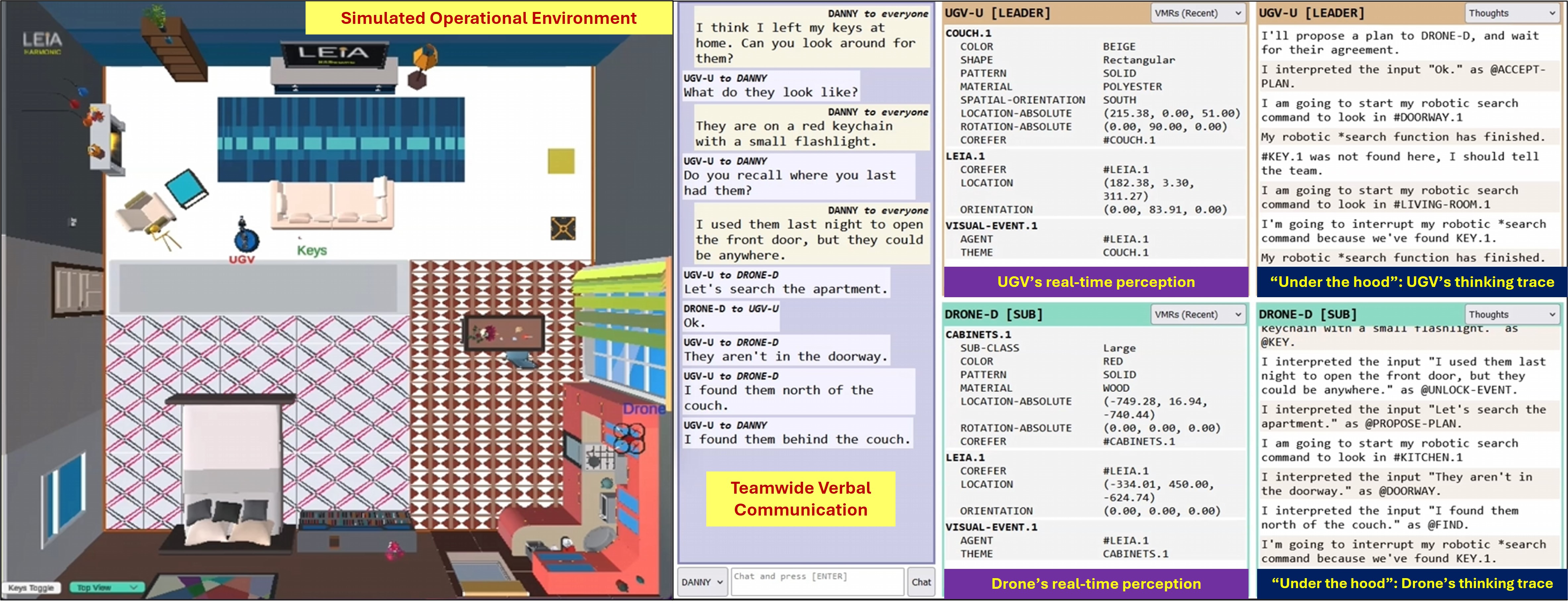}
    \caption{A snapshot of the simulation environment featuring a UGV and a drone searching for lost keys, as requested by a human named Danny. In the center is the team-wide verbal communication. To the right are under-the-hood panels that show real-time traces of thinking in the strategic layer, including interpreting visual information (VMRs) and reasoning (Thoughts).}
    \label{fig:simulation result}
    \vspace{-12pt}
\end{figure*}
HARMONIC is a dual control architecture consisting of a strategic (cognitive) layer for high-level decision-making and planning, a tactical (robot) layer for low-level robot control and execution, and a bidirectional interface to support communication between the layers, as shown in Figure \ref{fig:overview}. 

The strategic layer includes modules for attention management, perception interpretation, and utility-based and analogical decision-making, enhanced by metacognitive abilities supported by the microtheories of the OntoAgent cognitive architecture \cite{english2020ontoagent, nirenburg2011cognitive}. These modules prioritize the strategic goal and plan agenda, and select actions while monitoring their execution. Additional  team-oriented operations include natural language communication, explaining decisions, assessing decision confidence, and evaluating the trustworthiness of one's teammates \cite{nirenburg2024mutual}.

The tactical layer includes controllers, algorithms, and models responsible for decision-making at the robot control level. This involves processing sensor inputs and planning motor actions to execute high-level commands received from the strategic layer. For example, a command from the strategic layer to "pick up a screwdriver" requires the tactical component to identify the screwdriver, determine its position and orientation, compute the end-effector trajectory, and send control signals to the actuators. This layer also handles reactive responses, such as collision avoidance in robots, which are managed by a dedicated controller.

The strategic layer of the architecture relies on substantial knowledge bases: an ontological world model with 9,000 concepts; a lexicon that describes the meanings of $\sim$25,000 English word and phrase senses in terms of ontological concepts; and profiles of human and robotic agents that detail team roles, skills, preferences, and states. The current implementation of the strategic layer employs the OntoSem natural language analyzer \cite{mcshane2021linguistics}, an attention manager for goal prioritization, a deliberation module for decision-making, an action rendering module, a semantically-oriented text generator, and the DEKADE software environment, which all of the needs of the strategic layer -- Development, Evaluation, Knowledge Acquisition, and DEmonstration \cite{EnglishNirenburg2024}.

On the tactical side, we utilize Behavior Trees (BTs) \cite{colledanchise2018behavior,iovino2022survey}  for executing physical action plans and ensuring reactive control. BTs facilitate effective reactive robot control, they provide robust and modular representations for skills and low-level plans \cite{venkata2023kt,oruganti2023ikt} in dynamic environments, and they support HARMONIC's safety and operational needs. 

The strategic layer of HARMONIC implements Daniel Kahneman's \cite{kahneman2011thinking} System 2, or slow reasoning. The tactical layer implements System 1, or fast reasoning and reflexive action. At its core, the architecture facilitates dynamic scheduling and adaptation, enabling the system to adjust priorities and actions in real time. This is particularly important for capabilities such as responding to computational delays in the strategic component, handling contingencies, ensuring safety (as by avoiding collisions), and optimizing resources by engaging reactive planning algorithms in the tactical layer. 


\section{Initial Implementation}
Our initial implementation of the HARMONIC architecture involves a human-robot team carrying out simulated search and retrieval tasks in an apartment environment. The team includes a human and two HARMONIC-based robots -- a UGV and a drone, as shown in Figure \ref{fig:simulation result}. In this simulation, the robots are searching for a set of lost keys at the request of a human named Danny. They use dialogue to establish search parameters, select and execute a strategic-level plan, and coordinate their efforts. In the center is the team-wide verbal communication. To the right are under-the-hood panels that show real-time traces of thinking in the strategic layer, including interpreting visual information (VMRs) and reasoning (Thoughts). Each robot is equipped with customized BTs, which include higher-priority sub-trees to ensure the robots' safety and needs.

\section{CONCLUSIONS AND FUTURE WORK}

The development of the HARMONIC framework is part of a comprehensive research program to integrate advanced strategic, tactical, and infrastructure components of robots capable of effectively functioning as team members that can perform physical, mental, and hybrid actions. At this time we are planning to deploy the HARMONIC framework on advanced robotic systems and test them in real-time in the domain of ship maintenance tasks.  

\addtolength{\textheight}{-12cm}   








\bibliographystyle{IEEEtran}
\bibliography{references}

\end{document}